\documentclass[letterpaper, 10 pt, conference]{ieeeconf}  

\IEEEoverridecommandlockouts                              

\usepackage{graphics} 
\usepackage{epsfig} 
\usepackage{mathptmx} 
\usepackage{times} 
\usepackage{amsmath} 
\usepackage{amssymb}  
\usepackage{amsmath}
\usepackage{booktabs}   
\usepackage{tabularx}   
\makeatletter
\@ifundefined{labelindent}{}{\let\labelindent\relax}
\makeatother
\usepackage{enumitem}

\IEEEoverridecommandlockouts
\usepackage{capt-of}
\usepackage{capt-of}
\usepackage{booktabs}
\usepackage{xcolor}
\usepackage{pifont}
\usepackage{graphicx}
\usepackage{hyperref}

\newcommand{\capyes}{\textcolor[RGB]{0,128,70}{\ding{51}}}
\newcommand{\capno}{\textcolor[RGB]{195,40,40}{\ding{55}}}
\title{\LARGE \bf
From Social Reasoning to Embodied Interaction: An Agentic Framework for Social Robots
}

\author{%
\authorblockN{%
\makebox[\textwidth][c]{%
\begin{tabular}{c}
Ziyu Cheng$^{1,*}$, Yuewen Guo$^{1,*}$, Zhirui Liu$^{1,*}$,
Dong Zhang$^{1}$, Haotao Lu$^{1}$, Jingyi Yu$^{1}$,\\
Ye Shi$^{1,2}$, Jingya Wang$^{1,\dagger}$
\end{tabular}%
}%
}%
\authorblockA{%
\makebox[\textwidth][c]{%
\begin{tabular}{c}
\texttt{\{chengzy2023, guoyw2026, liuzhr2025, zhangdong2025,}\\
\texttt{luht2025, yujingyi, shiye, wangjingya\}@shanghaitech.edu.cn}
\end{tabular}%
}%
}%
\authorblockA{%
\makebox[\textwidth][c]{%
$^{1}$ShanghaiTech University \qquad $^{2}$InstAdapt
}%
}%
\thanks{$^{*}$These authors contributed equally to this work and are listed in alphabetical order by surname.}%
\thanks{$^{\dagger}$Corresponding author.}%
}

\IEEEaftertitletext{%
\vspace{-16pt}
  \begin{minipage}{\textwidth}
    \centering
    \includegraphics[width=\linewidth]{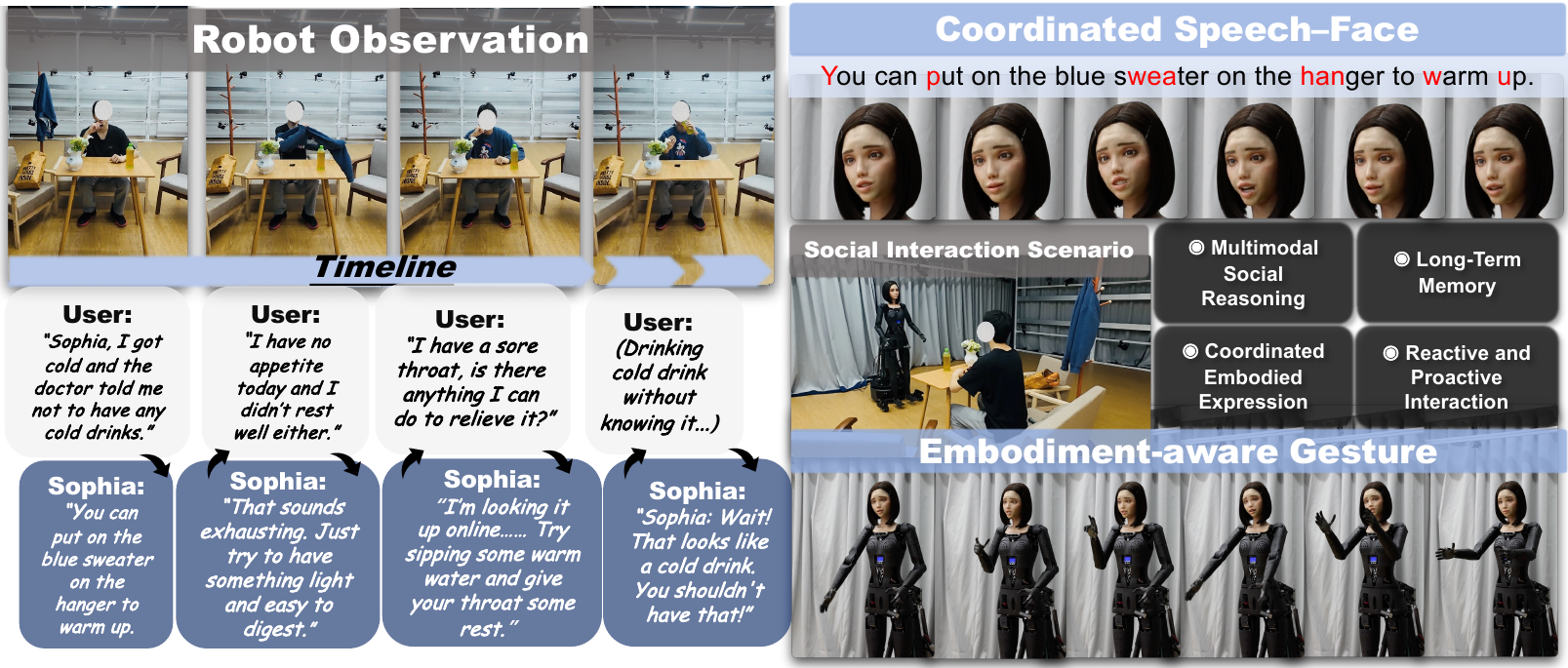}
    \captionof{figure}{\textbf{Illustration of \texttt{ARISE}.}
    Given robot observations and multi-round user interaction,
    Sophia integrates visual understanding, long-term interaction context,
    proactive engagement, robot-native gesture planning, and coordinated
    speech--face--gesture execution to produce context-aware embodied responses.}
    \label{fig:teaser}
    \vspace{0.5em}
  \end{minipage}%
}

\begin{document}

\maketitle
\thispagestyle{empty}
\pagestyle{empty}

\begin{abstract}
Natural face-to-face human--robot interaction requires a robot to understand an evolving social situation, decide when to engage, and express its intent through coordinated physical behavior. Yet existing approaches rarely close this loop: foundation-model agents provide increasingly capable multimodal reasoning and memory but remain largely disembodied, while expressive virtual agents do not face the physical constraints of real robots, and physical social robots typically address social reasoning and embodied expression only partially. We present \texttt{ARISE}, a unified framework that bridges \underline{A}gentic \underline{R}easoning and \underline{I}nteractive \underline{S}ocial \underline{E}mbodiment on the Sophia humanoid robot. \texttt{ARISE} integrates multimodal context understanding, long-term memory, and reactive and proactive interaction to determine when and what to communicate, and translates social intent into robot-native gestures coordinated with speech and mechanical facial expressions through streaming execution. Extensive evaluations on Sophia demonstrate strong perceived interaction quality, expressive and well-coordinated embodied behavior, and substantial latency reductions through streaming execution. These results highlight the importance of jointly reasoning about what to communicate, when to engage, and how to physically express social intent for natural interaction with humanoid robots. Project Page: \href{https://robosocial.github.io/}{https://robosocial.github.io/}

\end{abstract}

\begin{table*}[t]
\centering
\caption{Capability comparison with previous works.
\capyes: demonstrated; \capno: not reported.}
\label{tab:capability_comparison}
\footnotesize
\setlength{\tabcolsep}{4pt}
\renewcommand{\arraystretch}{1.18}
\begin{tabular*}{\textwidth}{@{\extracolsep{\fill}}l*{5}{c}@{}}
\toprule
Method &
\shortstack{Physical\\robot} &
\shortstack{Long-term\\memory} &
\shortstack{Proactive\\engagement} &
\shortstack{Facial\\expression} &
\shortstack{Co-speech\\gestures} \\
\midrule
EMAGE~\cite{liu2024emage}
& \capno & \capno & \capno & \capyes & \capyes \\
Social Agent~\cite{zhang2025social}
& \capno & \capno & \capno & \capyes & \capyes \\
MIBURI~\cite{mughal2026miburi}
& \capno & \capno & \capno & \capyes & \capyes \\
\midrule
Aly and Tapus~\cite{aly2015multimodal}
& \capyes & \capno & \capno & \capyes & \capyes \\
Attentive Support~\cite{tanneberg2024help}
& \capyes & \capno & \capyes & \capno & \capno \\
ExFace~\cite{zhang2025exface}
& \capyes & \capno & \capno & \capyes & \capno \\
Garello et al.~\cite{garello2025building}
& \capyes & \capyes & \capyes & \capyes & \capno \\
ProAct~\cite{zhang2026proact}
& \capyes & \capyes & \capyes & \capno & \capyes \\
\midrule
\textbf{Ours}
& \capyes & \capyes & \capyes & \capyes & \capyes \\
\bottomrule
\end{tabular*}
\vspace{-12pt}
\end{table*}

\section{INTRODUCTION}

Empowering embodied agents with social intelligence is central to realizing
general-purpose human--robot interaction. A capable social agent should not
only infer human intentions from multi-modal interactions and generate
appropriate responses, but also continuously perceive and reason about its
surroundings, maintain awareness over extended interactions, and proactively
engage with humans in real time~\cite{long2026seeing,tanneberg2024help}.
These capabilities become particularly compelling when embodied in humanoid
robots, where perception, reasoning, communication, and physical behavior are
tightly coupled in real-world interaction. Such physically grounded social
intelligence therefore represents a promising direction toward interactive
embodied agents that can naturally coexist and communicate with humans in
everyday environments.

Recent advances in foundation models have substantially expanded the
capabilities of digital social agents, enabling natural, context-aware
communication and long-horizon multimodal interaction
~\cite{long2026seeing}. Yet, such agents remain fundamentally disembodied,
lacking the nonverbal behaviors essential to social interaction. Video-based
avatars and 3D virtual humans have introduced expressive facial and full-body
behaviors, including human animation~\cite{lin2025omnihuman,sun2026streamavatar},
co-speech gesture generation~\cite{liu2024emage,mughal2026miburi}, and
real-time motion synthesis~\cite{ji2025towards,zhang2025social,jiang2026super},
but remain confined to virtual spaces. Transferring these capabilities from
digital embodiments to physical robots remains challenging: virtual motion is
largely unconstrained, whereas physical execution must respect limited degrees
of freedom, joint limits, and hardware-specific actuation constraints.
Physical social robots have begun to integrate foundation-model reasoning with
expressive behavior, supporting affect-aware interaction~\cite{li2025ugotme},
proactive assistance~\cite{tanneberg2024help}, interruption-aware
dialogue~\cite{cao2025interruption}, and expressive facial
behavior~\cite{li2024driving,zhang2025exface}. However, these capabilities
have largely been studied in isolation. A unified physical social agent that
can understand evolving social context, engage proactively, and coordinate verbal and nonverbal behavior remains largely unexplored.

To address this gap, we present a unified embodied social interaction system
deployed on the Sophia~\cite{hanson2020neuro} humanoid robot. At its core, a
multimodal social agent directly reasons over conversational audio and visual
observations, augmented with long-term memory, online knowledge retrieval, and
proactive interaction to support both reactive and self-initiated engagement.
An embodiment-aware gesture planner further translates response intent into
verified symbolic gesture programs composed of executable upper-body motion
primitives. To realize these high-level decisions as coherent physical
behavior, we introduce a streaming multimodal orchestration scheme that
synchronizes speech with audio-driven facial animation and schedules gestures
online using the audio--facial stream as a shared temporal reference. Together,
these components couple multimodal social reasoning with coordinated verbal
and nonverbal behavior, enabling context-aware and proactive face-to-face
interaction on a physical humanoid robot.

We summarize our core contributions as follows:
\begin{enumerate}[label=\arabic*), leftmargin=*, itemsep=0pt, topsep=2pt]

    \item We present \texttt{ARISE}, a unified framework for physical humanoid social interaction that combines multimodal social reasoning, long-term memory, and coordinated embodied expression, supporting both reactive and proactive engagement.

    \item We bridge high-level agentic social reasoning and low-level
    embodiment-aware gesture planning to translate social intent into executable robot behavior. Streaming Multimodal Orchestration further coordinates asynchronous speech, facial animation, and gesture streams for progressive embodied execution.
    
    \item Extensive evaluations on Sophia demonstrate strong context-aware social interaction, effective proactive engagement, and expressive, coordinated embodied behavior.

\end{enumerate}

\section{RELATED WORKS}

\subsection{Social Reasoning with Foundation Models}

Large models have expanded social interaction from dialogue generation toward multimodal understanding, memory, and contextual planning~\cite{zhang2023large}. Socratic Models~\cite{zeng2022socratic} and embodied multimodal models such as PaLM-E~\cite{driess2023palm} compose pretrained models through language to support multimodal reasoning and assistive dialogue. Generative Agents~\cite{park2023generative} integrate memory, reflection, and planning to produce sustained social behavior in simulated environments. M3-Agent~\cite{long2026seeing} further extends memory-based reasoning to visual and auditory experience through episodic and semantic memory. Together, these approaches provide increasingly capable foundations for understanding social context and formulating appropriate responses. However, social reasoning alone does not deliver the physical presence and expressive behavior needed for immersive interaction. Our framework bridges this gap by translating multimodal, memory-augmented reasoning into coordinated speech, facial expressions, and upper-body gestures on a physical humanoid.

\subsection{Co-speech Gesture Generation}

Co-speech gesture generation has advanced toward expressive, holistic animation of 3D digital humans. EMAGE~\cite{liu2024emage} integrates audio conditioning with masked gesture modeling to generate coordinated face, body, and hand motion. DiffSHEG~\cite{chen2024diffsheg} jointly models facial expressions and body gestures through diffusion, supporting synchronized motion generation for arbitrary-length speech. More recently, MIBURI~\cite{mughal2026miburi} enables online, causal generation of expressive gestures and facial expressions alongside streaming dialogue. These methods advance motion quality and responsiveness primarily in virtual human representations. However, transferring such expressiveness to physical robots requires accommodating fewer degrees of freedom and hardware-specific joint limits. We address this challenge by composing and verifying robot-native gesture programs to support semantic expressiveness and speech coordination within the robot's constrained motion repertoire.

\subsection{Multimodal Interaction in Social Robotics}

Social robotics has progressed from coordinated multimodal expression toward adaptive, reasoning-driven interaction. Early work by Aly and Tapus~\cite{aly2015multimodal} combines emotional speech, facial expressions, and speech-accompanying head and arm gestures in narrative interaction. More recent systems advance social decision-making: Attentive Support~\cite{tanneberg2024help} reasons about when and how to assist human groups, while Garello et al.~\cite{garello2025building} integrate LLM-based interaction management with knowledge-graph memory for adaptive tutoring. In parallel, speech-driven animatronic control and ExFace improve expressive facial actuation on physical robots~\cite{li2024driving,zhang2025exface}. ProAct~\cite{zhang2026proact} further connects memory-based proactive reasoning with streaming humanoid motion, bringing social planning and embodied behavior closer together.

However, these systems typically emphasize selected aspects of social reasoning or embodied expression, leaving their integration incomplete, as summarized in Table~\ref{tab:capability_comparison}. Our framework unifies memory-augmented reactive and proactive planning with coordinated speech, mechanical facial expressions, and verified upper-body gestures through a shared streaming pipeline, enabling context-aware, expressive, and responsive interaction on a physical humanoid.

\section{Method}
\label{sec:method}

\begin{figure*}[t]
    \centering
    \includegraphics[width=\textwidth]{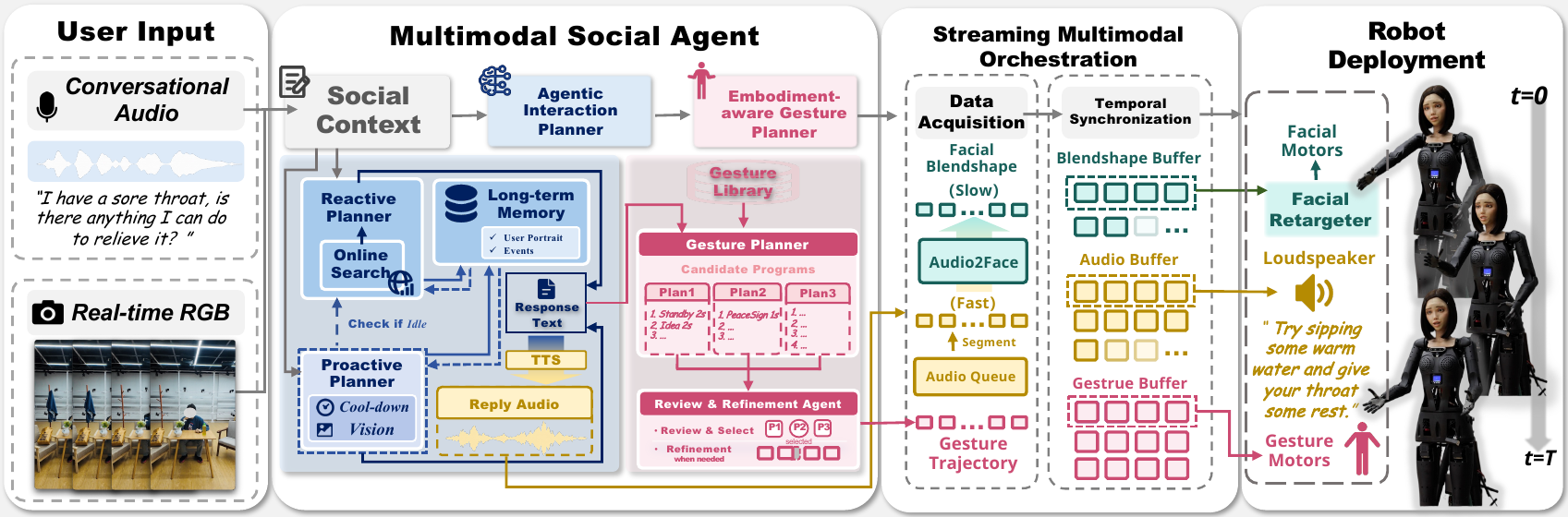}
    \caption{Overview of \texttt{ARISE}. Given conversational audio and real-time RGB observations, the \textit{Multimodal Social Agent} performs reactive and proactive planning with long-term memory and online search, generating a verbal response and symbolic gesture plan. \textit{Streaming Multimodal Orchestration} synchronizes speech, facial animation, and gestures for progressive execution, and \textit{Robot Deployment} maps the coordinated outputs to robot's actuators.}
    \label{fig:pipeline}
\end{figure*}

\subsection{Overview}

As illustrated in Fig.~\ref{fig:pipeline}, our framework maps ongoing
audio--visual interaction to coordinated verbal and nonverbal behaviors on
the Sophia~\cite{hanson2020neuro} human-like robot. Given conversational audio $A_t$ and RGB
observation $V_t$, the \textit{Multimodal Social Agent} generates a verbal
response and an executable symbolic gesture plan using multimodal reasoning,
long-term memory, online retrieval, and proactive interaction. The
\textit{Streaming Multimodal Orchestration} module converts these high-level
outputs into synchronized speech, facial animation, and temporally scheduled
gestures. The \textit{Robot Deployment} module finally maps the coordinated
signals to Sophia's actuators. This design separates semantic social planning
from temporal coordination and low-level execution.

\subsection{Multimodal Social Agent}

The \textit{Multimodal Social Agent} operates at the semantic level and produces a
verbal response $R_t$ and a symbolic gesture program $\hat{S}_t$. It contains
an \textit{Agentic Interaction Planner} for deciding what and when to
communicate and an \textit{Embodiment-aware Gesture Planner} for generating and refining executable upper-body behaviors \cite{yoon2019robots,huang2012robot,zhang2025social}.

\textbf{Agentic Interaction Planner.}
At interaction step $t$, the planner directly receives raw conversational
audio $A_t$ and the current RGB observation $V_t$ through an audio--visual
multimodal foundation model. We denote the current multimodal context by
$C_t=(A_t,V_t)$. The planner supports \textit{reactive} interactions triggered
by human speech and \textit{proactive} interactions initiated by the robot.

For a reactive interaction, a skill router determines whether long-term
memory and online search are needed:
\begin{equation}
D_t
=
\pi_{\mathrm{skill}}(C_t)
=
(d_t^{\mathrm{mem}},d_t^{\mathrm{search}}),
\label{eq:skill_routing}
\end{equation}
where $d_t^{\mathrm{mem}},d_t^{\mathrm{search}}\in\{0,1\}$ indicate whether
memory and online search are activated, respectively. The memory store $M_t$
preserves salient user information and interaction events across interactions.
When enabled, the current context is used to retrieve relevant entries
$\tilde{M}_t$. Online search is invoked when information beyond the model's
internal knowledge is required, with retrieved results filtered and summarized
into $\tilde{K}_t$. If a skill is not selected, its corresponding context is
left empty. The reactive response is then generated as
\begin{equation}
R_t^{\mathrm{react}}
=
f_{\theta}(C_t,\tilde{M}_t,\tilde{K}_t),
\label{eq:reactive_response}
\end{equation}
where $f_{\theta}$ denotes the multimodal response model.

Proactive interaction is evaluated independently of user-initiated dialogue.
To reduce unnecessary interruptions, evaluation is enabled only when the human
partner is silent, the \textit{Reactive Planner} and execution pipelines are idle, and a minimum cooldown interval has elapsed since the previous dialogue.
We denote this eligibility condition by $g_t\in\{0,1\}$.

Once eligible, a visual social estimator based on a VLM evaluates the current social context from the visual observation. The resulting assessment determines whether proactive interaction is appropriate:
\begin{equation}
z_t^{\mathrm{pro}}
=
f_{\psi}
\left(
V_t;P^{\mathrm{pro}}
\right),
\qquad
z_t^{\mathrm{pro}}
\in
\{\mathrm{initiate},\mathrm{wait}\},
\label{eq:proactive_assessment}
\end{equation}
where $P^{\mathrm{pro}}$ specifies the observable social cues and
decision criteria for proactive initiation. After a positive
decision, relevant long-term memories
$\tilde{M}_t^{\mathrm{pro}}$ are retrieved and supplied only to response
generation:
\begin{equation}
R_t^{\mathrm{pro}}
=
g_{\omega}
\left(
V_t,\tilde{M}_t^{\mathrm{pro}}
\right),
\qquad
d_t^{\mathrm{pro}}=1.
\label{eq:proactive_response}
\end{equation}
Thus, long-term memory affects the content of a proactive response but does
not influence whether the robot initiates the interaction.

\textbf{Embodiment-aware Gesture Planner.}
The gesture planner converts the social intent of $R_t$ into a symbolic
program executable on the robot. We define a semantic keyframe library
$K=\{k_i=(n_i,c_i,d_i)\}_{i=1}^{N}$, where $n_i$, $c_i$, and $d_i$ denote
the keyframe name, semantic category, and natural-language description,
respectively. The description $d_i$ provides a neutral semantic abstraction of
the motion, specifying what the keyframe visually conveys and in which social
contexts it is generally appropriate, without exposing low-level actuator
details to the planner. Each keyframe corresponds to a validated robot posture, while actuator-level mappings are handled during deployment. At a high level, the keyframes are organized into four semantic
categories: \textit{Neutral and Recovery}, \textit{Social and Affective},
\textit{Communicative and Presentational}, and \textit{Cognitive and
Attentive}.

Rather than generating a single gesture directly, the planner first proposes
$J$ candidate programs by reasoning over the response and the semantic keyframe
library:
\begin{equation}
\{S_t^{(j)}\}_{j=1}^{J}
=
h_{\phi}(R_t,K).
\label{eq:gesture_candidates}
\end{equation}
Each candidate is a keyframe sequence
$S_t^{(j)}=[(n_{j,1},\tau_{j,1}),\ldots,(n_{j,M_j},\tau_{j,M_j})]$,
where $n_{j,m}$ denotes the selected keyframe name and $\tau_{j,m}$ denotes its
nominal duration. Candidate construction follows semantic composition rules
encoded in the planner, such as pairing hand-shape primitives with a preceding
arm raise to preserve visibility, ending most gestures with a neutral standby
pose, and avoiding broad presenting motions when inconsistent with the response
intent.

The candidate set is then passed to a review-and-refinement agent:
\begin{equation}
\hat{S}_t
=
g_{\eta}\left(\{S_t^{(j)}\}_{j=1}^{J},R_t,K\right).
\label{eq:gesture_review_refinement}
\end{equation}
This stage shares the general motivation of prior work on grounding
language-model outputs in executable robot skills, such as SayCan~\cite{ahn2022can}, in that
unchecked language-model proposals should not be directly executed on
physical robots. The agent evaluates the candidates according to semantic fit, social
appropriateness, naturalness, timing compatibility, and embodiment feasibility.
When a candidate is already suitable, the agent selects it as the final symbolic
gesture plan. When minor issues are present, it may refine the sequence, for
example by replacing an unsuitable keyframe, adding a neutral ending, or
adjusting nominal durations to better match the spoken response. This stage
therefore serves both as a candidate-level reviewer and as a lightweight
refinement mechanism over the symbolic keyframe space.

The resulting plan
$\hat{S}_t=[(n_{t,1},\tau_{t,1}),\ldots,
(n_{t,\hat{M}_t},\tau_{t,\hat{M}_t})]$.
specifies \emph{what} gesture to perform using validated keyframe names from
the semantic library. Low-level actuator commands are not exposed to the
planner; they are resolved only during deployment through the predefined mapping
associated with each keyframe. The detailed temporal alignment between speech
and motion is handled by the subsequent coordination module.

\subsection{Streaming Multimodal Orchestration}

This module converts $R_t$ and $\hat{S}_t$ into a coordinated streaming
response. It first synchronizes speech and facial animation, then uses the
audio--facial stream as the reference clock for gesture scheduling.

The generated speech is streamed as temporally ordered audio chunks
$a_{t,r}$. Each chunk is sent in parallel to the audio buffer and Audio2Face,
which produces the corresponding facial blendshape chunk
\begin{equation}
f_{t,r}
=
f_{\mathrm{A2F}}(a_{t,r}).
\label{eq:audio2face}
\end{equation}
Because facial synthesis introduces additional latency, immediate audio
playback would cause the facial signal to lag behind speech. We therefore
enqueue an audio chunk only after its corresponding facial chunk becomes
available. Paired chunks $(a_{t,r},f_{t,r})$ enter a synchronization buffer,
and playback starts after $n_{\mathrm{buf}}$ complete pairs are accumulated.
The buffer is then drained in temporal order while subsequent pairs are
generated and enqueued online. This warm-up strategy trades an initial buffering delay for stable audio--facial synchrony without waiting for the complete
response.

The final gesture plan
$\hat{S}_t=[(n_{t,1},\tau_{t,1}),\ldots,(n_{t,M},\tau_{t,M})]$
is first constrained by a coarse estimate $\hat{T}_t$ of speech duration.
Its nominal duration is
\begin{equation}
T_t^{g}
=
\sum_{m=1}^{M}\tau_{t,m}.
\label{eq:gesture_duration}
\end{equation}
When $T_t^{g}$ exceeds the available speech-duration budget, keyframe
durations are adjusted before execution to prevent the gesture sequence from
substantially exceeding the spoken response. This planning-time adjustment
regulates the overall gesture density without imposing fine-grained
word-level alignment.

Runtime scheduling is anchored to the actual progress of streaming speech.
Let $\delta_{t,r}$ denote the duration of audio chunk $a_{t,r}$. After the
$r$-th chunk, the accumulated audio time is
\begin{equation}
T_{t,r}^{\mathrm{aud}}
=
\sum_{i=1}^{r}\delta_{t,i}.
\label{eq:audio_clock}
\end{equation}
Because each facial chunk is paired with the same audio chunk,
$T_{t,r}^{\mathrm{aud}}$ defines a shared audio--facial clock. The scheduler
maps the ordered keyframe names in $\hat{S}_t$ onto this clock. Each keyframe name
$n_{t,m}$ is assigned an execution interval
$I_{t,m}=[s_{t,m},s_{t,m}+\tau_{t,m}]$ using the current audio time and a
short look-ahead window to tolerate asynchronous generation and robot
playback. Gestures are therefore scheduled progressively as speech is
generated rather than after the full response becomes available.

\subsection{Robot Deployment}

Deploying \texttt{ARISE} on Sophia requires converting heterogeneous behavioral representations into robot-specific commands while preserving their temporal structure. Speech arrives as audio chunks, facial motion as blendshape sequences, and upper-body behavior as symbolic keyframes, each requiring a different physical realization. Audio is dispatched to the loudspeaker, while Audio2Face outputs are retargeted through ExFace [6] to Sophia’s mechanical facial actuators. Each gesture keyframe is resolved through a predefined mapping to a validated upper-body motor target. This representation-specific translation grounds behavior in the robot’s available actuation space without exposing hardware-level control parameters to the language-based planner.

At runtime, the deployment layer dispatches mapped commands according to the shared audio--facial clock established by Streaming Multimodal Orchestration. Audio playback and facial actuation follow the paired chunk sequence, while upper-body motor targets are issued according to their assigned intervals $I_{t,m}$, preserving the planned keyframe order and durations across modality-specific interfaces.

\section{EXPERIMENTS}
To validate the performance of the framework and assess its practical
applicability to embodied human--robot interaction, we investigate the following research questions:

\noindent\textbf{Q1.} How effectively does the framework support sustained and context-aware multimodal social interaction?

\noindent\textbf{Q2.} How effectively does the framework generate expressive robot behaviors that are semantically and temporally aligned with the interaction content?

\noindent\textbf{Q3.}  How effectively do the proposed engineering strategies mitigate latency accumulation across cascaded multimodal processing?

\subsection{Experimental Setup}
\textbf{Hardware Setup.}
We conduct our experiments on the Sophia~\cite{hanson2020neuro} humanoid robot developed by Hanson Robotics. An Intel RealSense RGB camera is mounted between Sophia's head and chest to capture visual observations, and a Philips DLM3545C Bluetooth microphone captures speech. Our framework runs on a host computer equipped with an NVIDIA RTX 3090 GPU, which communicates with Sophia through the robot's native control interface to deliver speech, facial expressions, and upper-body gestures.

\textbf{Evaluation Metrics.}
All ratings were collected using a five-point Likert scale, where 1 denotes
``strongly disagree'' and 5 denotes ``strongly agree.'' We used different
rating criteria for the two user studies.

For the \textit{Multi-round Interaction} experiment, participants evaluated each interaction session using four criteria:
\begin{itemize}
   \item \textit{Overall Interaction Quality} (OIQ): To what extent do the robot's verbal and non-verbal behaviors together form a coherent and satisfactory social interaction?
   \item \textit{Contextual Appropriateness} (CA): To what extent is the robot's response appropriate given the user's utterance, visible environment, and current interaction context?
   \item \textit{Conversational Continuity} (CC): To what extent does the robot's response remain consistent with information and events established earlier in the interaction?
   \item \textit{Interaction Engagement} (IE): To what extent did the interaction maintain your interest and attention?
\end{itemize}

For the \textit{Gesture Generation} experiment, participants evaluated each interaction session using four criteria:
\begin{itemize}
   \item \textit{Motion Naturalness}: To what extent do the robot's movements appear smooth, physically
    plausible, and natural?
   \item \textit{Motion Expressiveness}: To what extent do the robot's movements convey the intended social or communicative meaning?
   \item \textit{Speech-Motion Coordination}: To what extent are the robot's movements temporally and semantically coordinated with its speech?
   \item \textit{Overall Quality}: Considering both the robot's speech and movements, how would you rate
    the overall quality of this response?
\end{itemize}

\textbf{User Study.}
We recruited 20 participants who had no prior knowledge of our system for each of Q1 and Q2. Each participant completed the full evaluation for all configurations within the corresponding experiment.

For Q1, participants were provided with a predefined dialogue script to ensure
comparable interaction content, while retaining the freedom to ask natural
follow-up questions based on the robot's responses. For Q2, participants were
provided with predefined dialogue scripts and were asked to follow the scripts
while focusing on the robot's verbal and gestural responses.

After each interaction, participants rated the robot according to the evaluation
metrics associated with the corresponding study. The order of configurations was
independently randomized for each participant. Participants were not allowed to see or interact with the robot before the experiment, and not informed of
the configuration used in each interaction or whether it corresponded to the
baseline or an ablation. Before each interaction, the robot was reset to the same predefined initial state. 

\textbf{Statistical Analysis.}
For each configuration, mean and standard deviation were computed across the
20 participant ratings. We applied two-sided paired Wilcoxon signed-rank tests
~\cite{wilcoxon1945individual} to participant-level scores. $P$-values were
adjusted using the Holm procedure~\cite{holm1979simple}.  Significance was denoted by $^{*}$ for $P_{\mathrm{adj}}<0.05$, $^{**}$ for $P_{\mathrm{adj}}<0.01$, and $^{***}$ for $P_{\mathrm{adj}}<0.001$.

\subsection{Baselines and Ablation}
To assess the contributions of individual components to interaction-response
and embodied-gesture quality, we evaluated the Full System together with the
following controlled ablation settings.

For Q1, we compare four configurations: \textbf{Full System}, which enables
visual input, long-term memory, and proactive interaction; \textbf{w/o Visual
Input}, which removes the current RGB observation; \textbf{w/o Long-Term
Memory}, which disables memory retrieval and update; and \textbf{w/o Proactive
Interaction}, which disables proactive behaviors while retaining user-initiated
response generation.

For Q2, we compare four configurations of our \textit{Embodiment-Aware Gesture
Generation} framework. \textbf{Full System} denotes the complete configuration. \textbf{Single Candidate} changes the number of generated
candidates from \(J=3\) to \(J=1\). \textbf{w/o Review-and-Refinement Agent}
removes the review-and-refinement stage and directly generates and executes a
single gesture sequence. We also include EMAGE~\cite{liu2024emage} as an external speech-driven gesture
baseline. EMAGE generates human-body motion in the SMPL-X representation. We
retain only its upper-body motion and project it into Sophia's joint space.
Since Sophia's joints are modeled as single-degree-of-freedom actuators, this
projection reduces the corresponding SMPL-X motion to one-dimensional joint
commands. Before execution, a safety check clips any command that exceeds the
allowable joint range of the robot.

\subsection{Implementation Details and Results}
\textbf{Implementation Details.}
In the experiment, the \textit{Reactive Planner}, the maintenance \textit{Long-term Memory}, the visual social estimator and the memory retrieved response generation of \textit{Proactive Planner} in our framework are implemented using Doubao-Seed-2.1-Pro-260628. The Audio2Face module employs the NVIDIA Audio2Face-3D v2.3 model~\cite{chung2025audio2face}, and the facial retargeter employs the ExFace~\cite{zhang2025exface} model. 
For gesture generation, the language-agent modules are implemented using
DeepSeek-V4. The
final gesture sequence is compiled into robot actuator commands
during deployment. In the Full System, the planner generates \(J=3\) candidate
symbolic gesture programs for each response. The full executable library contains 37
keyframes and consists of 2 \textit{Neutral and Recovery} keyframes, 9
\textit{Social and Affective} keyframes, 24
\textit{Communicative and Presentational} keyframes, and 2
\textit{Cognitive and Attentive} keyframes. Each keyframe is associated with a
validated robot posture and a semantic description used by the gesture
planner. The review-and-refinement agent
scores these candidates, selects the best one, and refines it when necessary.
\begin{figure*}[!t]
    \centering
    \setlength{\abovecaptionskip}{3pt}

    \includegraphics[width=\linewidth]{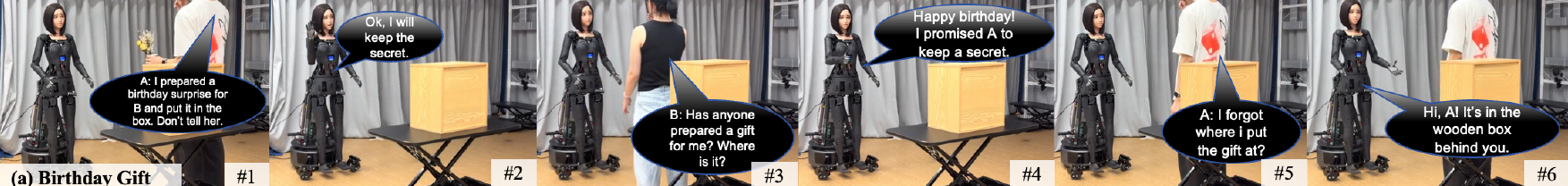}\\
    \vspace*{0.3pt}

    \includegraphics[width=\linewidth]{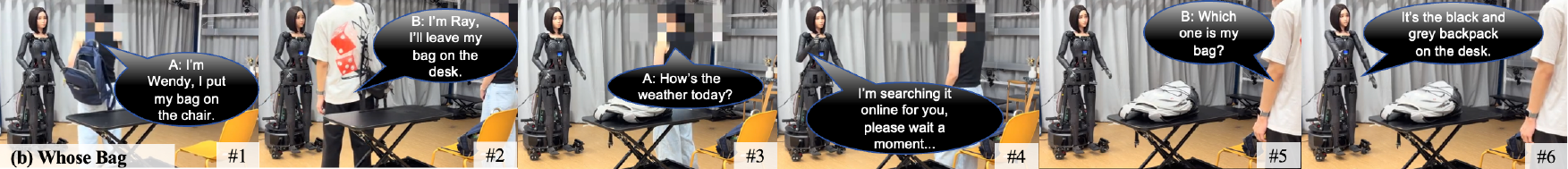}\\
    \vspace*{0.3pt}

    \includegraphics[width=\linewidth]{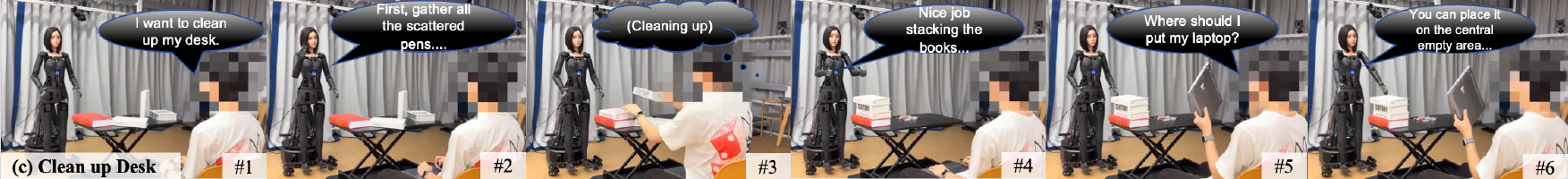}

    \caption{\textbf{Multi‑round Interaction Sequences}
(a) \textbf{Birthday Gift}: Sophia preserves confidential information for individual users and remembers the gift’s specific location.
(b) \textbf{Whose Bag}: Leveraging visual perception to distinguish different users and backpacks, Sophia recalls object‑user associations and answers identity‑related queries.
(c) \textbf{Clean up Desk}: Sophia can offer appropriate contextual guidance for object placement, and give positive feedback proactively after the user finishes the task.}
    \label{fig:q1_examples}
\end{figure*}

 \begin{table}[t]
  \centering
  \scriptsize
  \setlength{\tabcolsep}{2.5pt}
  \renewcommand{\arraystretch}{1.12}
 \caption{Results for Multi-Round Social Interaction.}
  \label{tab:q1_results}
  \resizebox{\columnwidth}{!}{%
  \begin{tabular}{@{}lcccc@{}}
  \toprule
  \ \textbf{Method}
  & \shortstack{\textbf{OIQ}\textbf{($\uparrow$)}}
  & \shortstack{\textbf{CA}\textbf{($\uparrow$)}}
  & \shortstack{\textbf{CC}\textbf{($\uparrow$)}}
  & \shortstack{\textbf{IE}\textbf{($\uparrow$)}} \\
  \midrule
  Full System
  & \textbf{4.38 $\pm$ 0.41}
  & \textbf{4.34 $\pm$ 0.43}
  & \textbf{4.25 $\pm$ 0.46}
  & \textbf{4.31 $\pm$ 0.44}\\
  w/o Visual Input
  & 3.72 $\pm$ 0.58\textsuperscript{***}
  & 2.90 $\pm$ 0.69\textsuperscript{***}
  & 4.08 $\pm$ 0.50
  & 4.06 $\pm$ 0.52 \\
  w/o Long-Term Memory
  & 3.54 $\pm$ 0.61\textsuperscript{***}
  & 3.78 $\pm$ 0.57\textsuperscript{**}
  & 3.29 $\pm$ 0.68\textsuperscript{***}
  & 3.62 $\pm$ 0.64\textsuperscript{**}\\
  w/o Proactive Interaction
  & 3.76 $\pm$ 0.51\textsuperscript{**}
  & 4.23 $\pm$ 0.44
  & 4.16 $\pm$ 0.47
  & 3.12 $\pm$ 0.71\textsuperscript{***}\\
  \bottomrule
  \end{tabular}%
  }
  \end{table}

\textbf{Multi-round Interaction (Q1)}
Fig.~\ref{fig:q1_examples} presents some characteristic interaction sequences, including online searching, memory-supported conversational continuity and visually grounded proactive assistance. As shown in Table~\ref{tab:q1_results}, the Full System achieves the best overall performance across all criteria. Removing visual input causes CA to decrease sharply, showing that visual information is essential for understanding interaction situation. Removing long-term memory lowers all four scores. This shows that interaction understanding, especially proactive behavior, depends substantially on contextual information. Underlying language model has some basic memory capability, but cannot maintain the persistent and precise contextual connections provided by the long-term memory module. Without proactive interaction, IE and OIQ decrease, while CA and CC remain close to the Full System. This 
shows that proactive functions, such as reminders and corrections, primarily improve user engagement and the practical value of the interaction.

\begin{figure}[!t]
    \centering
    \setlength{\abovecaptionskip}{3pt}

    \includegraphics[width=\linewidth]{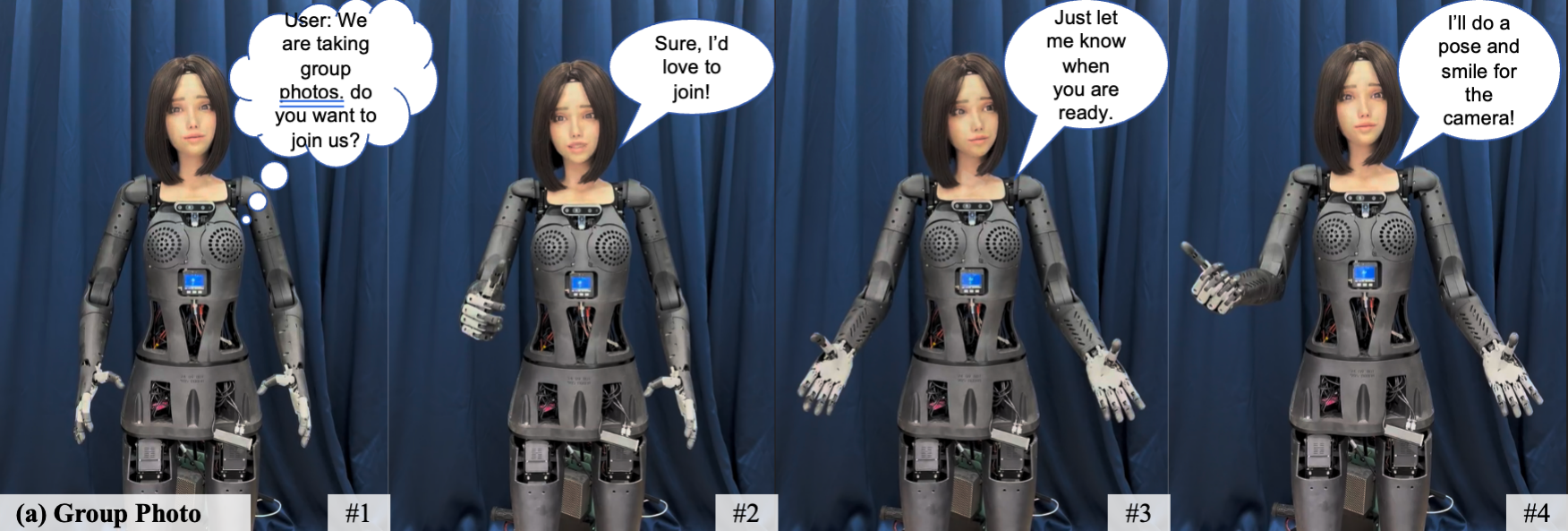}\\
    \vspace*{-0.4pt}

    \includegraphics[width=\linewidth]{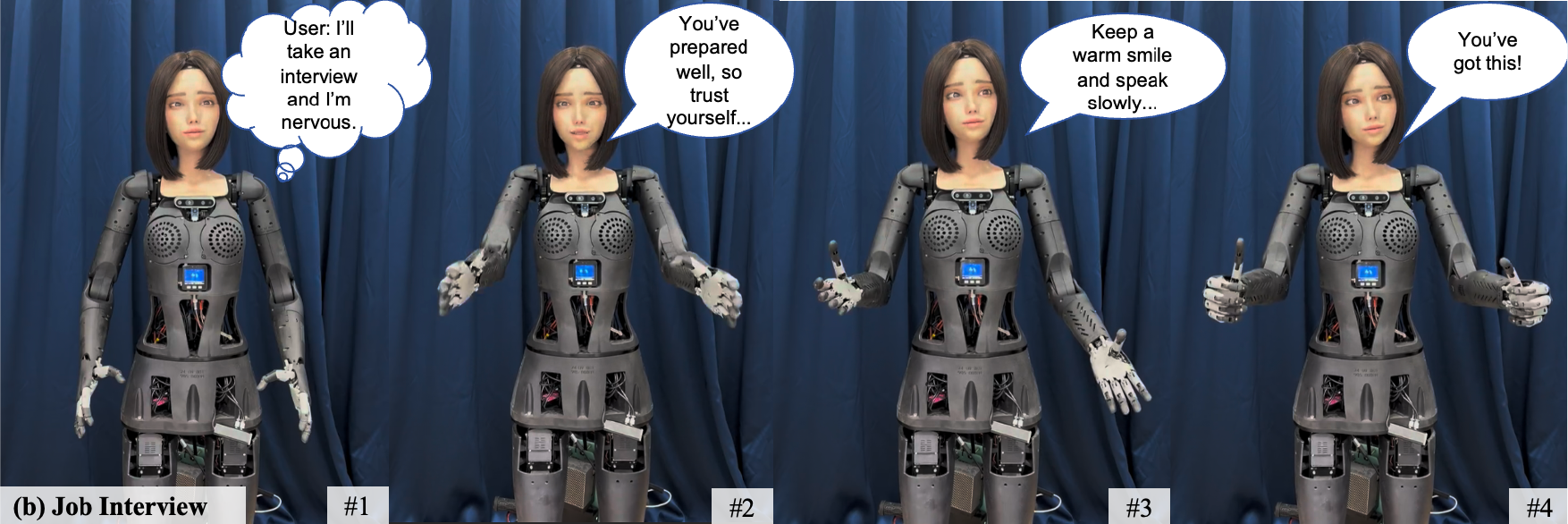}

    \caption{\textbf{Gesture Generation Sequences}
    (a) \textbf{Group Photo}: The robot actively responds to social invitations and cooperates for group‑photo scenarios.
    (b) \textbf{Job Interview}: Detecting user nervousness, Sophia delivers comforting verbal feedback and appropriate expressive gestures.}
    \label{fig:q2_examples}
\end{figure}

\begin{table}[t]
\raggedright
\scriptsize
\setlength{\tabcolsep}{2.5pt}
\renewcommand{\arraystretch}{1.05}

\caption{Results for the Gesture-Generation Experiment.}
\label{tab:q2_gesture_results}

\begin{tabular*}{\columnwidth}{@{\extracolsep{\fill}}lcccc}

\hline
\textbf{Method}
& \shortstack{\textbf{Motion} \\ \textbf{Natural.}}
& \shortstack{\textbf{Motion} \\ \textbf{Express.}}
& \shortstack{\textbf{Speech-Motion} \\ \textbf{Coord.}}
& \shortstack{\textbf{Overall} \\ \textbf{Quality}} \\
\hline

Full System
& \textbf{3.92 $\pm$ 0.34}
& \textbf{4.22 $\pm$ 0.34}
& \textbf{4.03 $\pm$ 0.60}
& \textbf{4.09 $\pm$ 0.26} \\

Single Candidate
& 3.63 $\pm$ 0.41\textsuperscript{**}
& 3.88 $\pm$ 0.43\textsuperscript{*}
& 3.79 $\pm$ 0.41
& 3.92 $\pm$ 0.29 \\

w/o R-R Agent
& 3.00 $\pm$ 0.39\textsuperscript{***}
& 3.35 $\pm$ 0.46\textsuperscript{***}
& 3.21 $\pm$ 0.44\textsuperscript{**}
& 3.58 $\pm$ 0.24\textsuperscript{***} \\

EMAGE
& 3.37 $\pm$ 0.27\textsuperscript{***}
& 3.67 $\pm$ 0.38\textsuperscript{***}
& 3.89 $\pm$ 0.33
& 3.98 $\pm$ 0.25 \\

\hline
\end{tabular*}
\end{table}

\textbf{Gesture Generation (Q2).}
Figure~\ref{fig:q2_examples} shows representative social‑aware interaction examples. As shown in Table~\ref{tab:q2_gesture_results}, multi-candidate generation and review-and-refinement
provide complementary benefits for embodied gesture generation. The performance
gap between the Full System and Single Candidate indicates that generating
multiple candidates gives the reviewer a broader range of gesture programs to
evaluate, leading to more suitable embodied responses. Meanwhile, the
performance of Single Candidate relative to w/o Review-and-Refinement Agent
shows that reviewing and refining the generated program remains beneficial even
when only one candidate is available. EMAGE achieves competitive performance,
but its lower motion naturalness and motion expressiveness suggest that
embodiment-aware planning with executable gesture library is better
suited to the physical robot.

\subsection{More Experiments}
     
\begin{table}[t]
    \caption{Latency definitions.}
    \label{tab:latency_definitions}
    \centering
    \footnotesize
    \renewcommand{\arraystretch}{1.02}
    \begin{tabularx}{\columnwidth}{@{}>{\centering\arraybackslash}p{0.15\columnwidth}@{\hspace{6pt}}X@{}}
      \toprule
      \textbf{Latencies} & \textbf{Definition} \\
      \midrule
      $L_{\mathrm{MSA}}$ & The latency of \textit{Multimodal Social Agent}. \\
      $L_{\mathrm{ES}}$ & The latency of embodied synthesis. \\
      $L_{\mathrm{robot}}$ & The latency of \textit{Robot Deployment}. \\
      $L_{\mathrm{AF}}$ & The end-to-end latency from the end of user's speech to the beginning of audio-facial playback. \\
      \bottomrule
    \end{tabularx}
\end{table}

\begin{table}[t]
  \caption{Results for System latency evaluation. Values are reported as mean.}
  \label{tab:system_latency}
  \centering
  \scriptsize
  \renewcommand{\arraystretch}{1.15}
  \setlength{\tabcolsep}{4pt}
  \begin{tabularx}{\linewidth}{@{}X@{\hspace{3pt}}cccc@{}}
      \toprule
      \textbf{Method}
      & $L_{\mathrm{MSA}}$ ($\downarrow$)
      & $L_{\mathrm{ES}}$ ($\downarrow$)
      & $L_{\mathrm{robot}}$ ($\downarrow$)
      & $L_{\mathrm{AF}}$ ($\downarrow$)\\
      \midrule
      Streaming Multimodal Orchestration (Ours)
      & 3.306
      & \textbf{1.906}
      & 0.003
      & \textbf{5.214} \\
      Serial Embodiment Processing
      & 3.227
      & 10.621
      & 0.003
      & 13.851 \\
      \bottomrule
  \end{tabularx}
\end{table}

\noindent\textbf{System Latency Evaluation (Q3)}
For Q3, we compare the latency of our proposed \textit{Streaming Multimodal Orchestration} with the \textit{Serial Embodiment Processing} baseline, which generates reply audio and facial animations serially. The benchmark uses 40 short dialogue segments sampled from the participant interactions. Table~\ref{tab:latency_definitions} summarizes the latency definitions. As shown in Table~\ref{tab:system_latency}, the latency of \textit{Serial Embodiment Processing} is far beyond the responsiveness required for natural HRI. In contrast, our proposed \textit{Streaming Multimodal Orchestration} substantially reduces the system delay.

\noindent\textbf{Effect of Executable Gesture Library Size.}
We further evaluate the effect of gesture library size using five keyframe
libraries containing 8, 15, 30, 37, and 45 keyframes, respectively. The
following category counts are reported in the order of \textit{Neutral and
Recovery}, \textit{Social and Affective}, \textit{Communicative and
Presentational}, and \textit{Cognitive and Attentive}. The 8-keyframe library
contains $(1,4,2,1)$ keyframes across these four categories, respectively. The
15-keyframe library expands these counts to $(2,5,7,1)$, while the
30-keyframe library further expands them to $(2,8,18,2)$. The 37-keyframe
library used in the main experiments contains $(2,9,24,2)$ keyframes. Finally,
the 45-keyframe library further adds \textit{Social and Affective} and
\textit{Communicative and Presentational} motions, resulting in
$(2,12,29,2)$ keyframes across the four categories. All other components,
including gesture planning, review-and-refinement, and streaming
orchestration, remain unchanged.

\begin{figure}[t]
\centering
\includegraphics[width=\columnwidth]{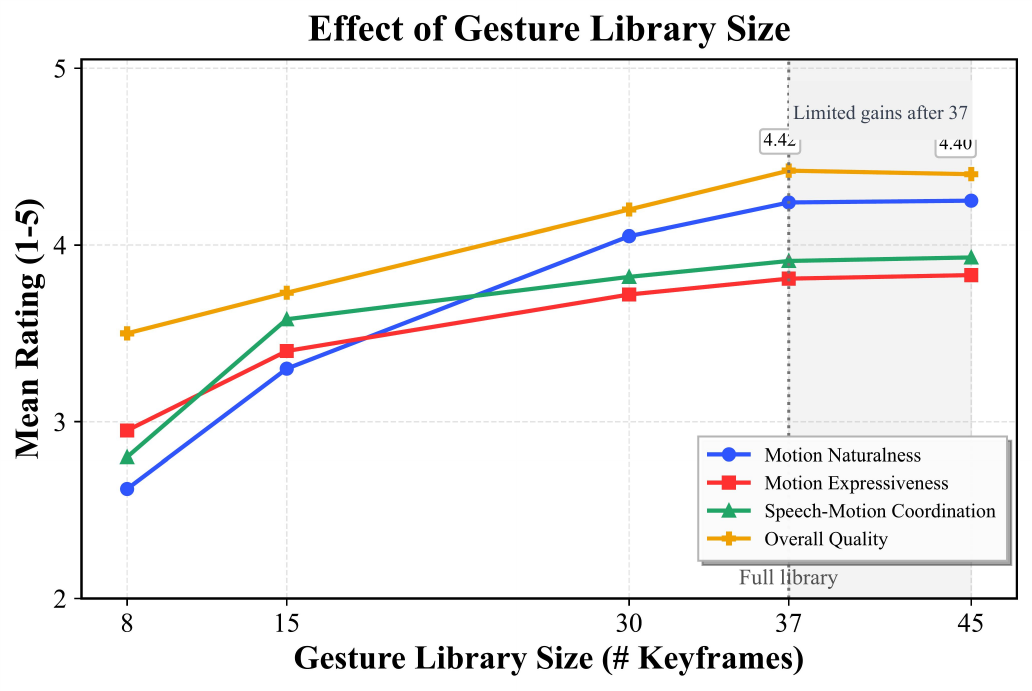}
\caption{Effect of gesture library size on user-rated motion quality. Library
size is measured by the number of semantic keyframes. The scores increase
substantially as the library grows from 8 to 37 keyframes, while the gain
becomes marginal when further expanded to 45 keyframes.}
\label{fig:gesture_library_size}
\end{figure}

As shown in Fig.~\ref{fig:gesture_library_size}, increasing the keyframe
library from 8 to 15, 30, and 37 keyframes leads to clear gains in motion naturalness, motion expressiveness, speech-motion coordination,
and overall quality. Beyond 37 keyframes, the effect of
further expanding the library becomes substantially weaker, showing only marginal changes and no consistent additional improvement.

\section{Limitations}
The current cascaded architecture still limits sub-second conversational responsiveness, with most latency arising from multimodal-agent inference and streaming facial synthesis rather than robot execution. 
Fine-grained facial expressiveness also remains limited on
the current hardware platform. Mechanical facial actuation constrains
the subtle motions that can be physically realized, intro-
ducing platform-dependent limits distinct from the proposed
social planning and orchestration mechanisms.
Future work will explore lightweight models specialized for social interaction and robot-native facial synthesis to further reduce response latency and improve embodied responsiveness.

\section{Conclusion}

We presented \texttt{ARISE}, a unified framework that connects multimodal social reasoning with coordinated embodied interaction on the Sophia humanoid. By integrating memory-augmented reactive and proactive interaction, robot-native gesture planning, and streaming speech–face–gesture orchestration, \texttt{ARISE} improves perceived interaction and gesture quality while reducing latency relative to serial embodiment processing. These results highlight the value of jointly considering what and when to communicate and how to express that intent through the robot’s physical capabilities.



\bibliographystyle{IEEEtran}
\bibliography{reference}

@inproceedings{long2026seeing,
  title={Seeing, listening, remembering, and reasoning: A multimodal agent with long-term memory},
  author={Long, Lin and He, Yichen and Ye, Wentao and Pan, Yiyuan and Lin, Yuan and Li, Hang and Zhao, Junbo and Li, Wei},
  booktitle={International Conference on Learning Representations},
  volume={2026},
  pages={146197--146246},
  year={2026}
}

@inproceedings{tanneberg2024help,
  title={To help or not to help: Llm-based attentive support for human-robot group interactions},
  author={Tanneberg, Daniel and Ocker, Felix and Hasler, Stephan and Deigmoeller, Joerg and Belardinelli, Anna and Wang, Chao and Wersing, Heiko and Sendhoff, Bernhard and Gienger, Michael},
  booktitle={2024 IEEE/RSJ International Conference on Intelligent Robots and Systems (IROS)},
  pages={9130--9137},
  year={2024},
  organization={IEEE}
}

@inproceedings{li2025ugotme,
  title={Ugotme: An embodied system for affective human-robot interaction},
  author={Li, Peizhen and Cao, Longbing and Wu, Xiao-Ming and Yu, Xiaohan and Yang, Runze},
  booktitle={2025 IEEE International Conference on Robotics and Automation (ICRA)},
  pages={5542--5548},
  year={2025},
  organization={IEEE}
}

@inproceedings{lin2025omnihuman,
  title={Omnihuman-1: Rethinking the scaling-up of one-stage conditioned human animation models},
  author={Lin, Gaojie and Jiang, Jianwen and Yang, Jiaqi and Zheng, Zerong and Liang, Chao and Zhang, Yuan and Liu, Jingtuo},
  booktitle={2025 IEEE/CVF International Conference on Computer Vision (ICCV)},
  pages={13848--13858},
  year={2025},
  organization={IEEE}
}

@inproceedings{sun2026streamavatar,
  title={Streamavatar: Streaming diffusion models for real-time interactive human avatars},
  author={Sun, Zhiyao and Peng, Ziqiao and Ma, Yifeng and Chen, Yi and Zhou, Zhengguang and Zhou, Zixiang and Zhang, Guozhen and Zhang, Youliang and Zhou, Yuan and Lu, Qinglin and others},
  booktitle={Proceedings of the IEEE/CVF Conference on Computer Vision and Pattern Recognition},
  pages={10887--10897},
  year={2026}
}

@article{mughal2026miburi,
  title={Miburi: Towards expressive interactive gesture synthesis},
  author={Mughal, M Hamza and Dabral, Rishabh and Demberg, Vera and Theobalt, Christian},
  journal={arXiv preprint arXiv:2603.03282},
  year={2026}
}

@inproceedings{zhang2025exface,
  title={Exface: Expressive facial control for humanoid robots with diffusion transformers and bootstrap training},
  author={Zhang, Dong and Peng, Jingwei and Jiao, Yuyang and Gu, Jiayuan and Yu, Jingyi and Chen, Jiahao},
  booktitle={2025 IEEE/RSJ International Conference on Intelligent Robots and Systems (IROS)},
  pages={20263--20268},
  year={2025},
  organization={IEEE}
}

@article{cao2025interruption,
  title={Interruption handling for conversational robots},
  author={Cao, Shiye and Moon, Jiwon and Mahmood, Amama and Antony, Victor Nikhil and Xiao, Ziang and Liu, Anqi and Huang, Chien-Ming},
  journal={arXiv preprint arXiv:2501.01568},
  year={2025}
}

@inproceedings{li2024driving,
  title={Driving animatronic robot facial expression from speech},
  author={Li, Boren and Li, Hang and Liu, Hangxin},
  booktitle={2024 IEEE/RSJ International Conference on Intelligent Robots and Systems (IROS)},
  pages={7012--7019},
  year={2024},
  organization={IEEE}
}

@inproceedings{ji2025towards,
  title={Towards immersive human-x interaction: A real-time framework for physically plausible motion synthesis},
  author={Ji, Kaiyang and Shi, Ye and Jin, Zichen and Chen, Kangyi and Xu, Lan and Ma, Yuexin and Yu, Jingyi and Wang, Jingya},
  booktitle={2025 IEEE/CVF International Conference on Computer Vision (ICCV)},
  pages={10173--10183},
  year={2025},
  organization={IEEE}
}

@inproceedings{zhang2025social,
  title={Social agent: Mastering dyadic nonverbal behavior generation via conversational llm agents},
  author={Zhang, Zeyi and Zhou, Yanju and Yao, Heyuan and Ao, Tenglong and Zhan, Xiaohang and Liu, Libin},
  booktitle={Proceedings of the SIGGRAPH Asia 2025 Conference Papers},
  pages={1--12},
  year={2025}
}

@article{zhang2026proact,
  title={ProAct: A Dual-System Framework for Proactive Embodied Social Agents},
  author={Zhang, Zeyi and Kang, Zixi and Zhao, Ruijie and Feng, Yusen and Jiang, Biao and Liu, Libin},
  journal={arXiv preprint arXiv:2602.14048},
  year={2026}
}

@article{zeng2022socratic,
  title={Socratic models: Composing zero-shot multimodal reasoning with language},
  author={Zeng, Andy and Attarian, Maria and Ichter, Brian and Choromanski, Krzysztof and Wong, Adrian and Welker, Stefan and Tombari, Federico and Purohit, Aveek and Ryoo, Michael and Sindhwani, Vikas and others},
  journal={arXiv preprint arXiv:2204.00598},
  year={2022}
}

@inproceedings{park2023generative,
  title={Generative agents: Interactive simulacra of human behavior},
  author={Park, Joon Sung and O'Brien, Joseph and Cai, Carrie Jun and Morris, Meredith Ringel and Liang, Percy and Bernstein, Michael S},
  booktitle={Proceedings of the 36th annual acm symposium on user interface software and technology},
  pages={1--22},
  year={2023}
}

@inproceedings{yoon2019robots,
  title={Robots learn social skills: End-to-end learning of co-speech gesture generation for humanoid robots},
  author={Yoon, Youngwoo and Ko, Woo-Ri and Jang, Minsu and Lee, Jaeyeon and Kim, Jaehong and Lee, Geehyuk},
  booktitle={2019 International conference on robotics and automation (ICRA)},
  pages={4303--4309},
  year={2019},
  organization={IEEE}
}

@inproceedings{liu2024emage,
  title={Emage: Towards unified holistic co-speech gesture generation via expressive masked audio gesture modeling},
  author={Liu, Haiyang and Zhu, Zihao and Becherini, Giorgio and Peng, Yichen and Su, Mingyang and Zhou, You and Zhe, Xuefei and Iwamoto, Naoya and Zheng, Bo and Black, Michael J},
  booktitle={2024 IEEE/CVF Conference on Computer Vision and Pattern Recognition (CVPR)},
  pages={1144--1154},
  year={2024},
  organization={IEEE}
}

@inproceedings{chen2024diffsheg,
  title={Diffsheg: A diffusion-based approach for real-time speech-driven holistic 3d expression and gesture generation},
  author={Chen, Junming and Liu, Yunfei and Wang, Jianan and Zeng, Ailing and Li, Yu and Chen, Qifeng},
  booktitle={2024 IEEE/CVF Conference on Computer Vision and Pattern Recognition (CVPR)},
  pages={7352--7361},
  year={2024},
  organization={IEEE}
}

@inproceedings{huang2012robot,
  title={Robot behavior toolkit: generating effective social behaviors for robots},
  author={Huang, Chien-Ming and Mutlu, Bilge},
  booktitle={Proceedings of the seventh annual ACM/IEEE international conference on Human-Robot Interaction},
  pages={25--32},
  year={2012}
}

@article{hanson2020neuro,
  title={A neuro-symbolic humanlike arm controller for sophia the robot},
  author={Hanson, David and Imran, Alishba and Vellanki, Abhinandan and Kanagaraj, Sanjeew},
  journal={arXiv preprint arXiv:2010.13983},
  year={2020}
}

@article{chung2025audio2face,
  title={Audio2face-3d: Audio-driven realistic facial animation for digital avatars},
  author={Chung, Chaeyeon and Fedorov, Ilya and Huang, Michael and Karmanov, Aleksey and Korobchenko, Dmitry and Ribera, Roger and Seol, Yeongho and others},
  journal={arXiv preprint arXiv:2508.16401},
  year={2025}
}

@article{jiang2026super,
  title={Super Star: Towards Streaming Real-time Interactive Agents for Digital Humans},
  author={Jiang, Wentao and Xie, Youchen and Fan, Haidi and Chen, Yajing and Wang, Xin and Shi, Ye and Wang, Jingya},
  journal={arXiv preprint arXiv:2608.24909},
  year={2026}
}

@inproceedings{garello2025building,
  title={Building knowledge from interactions: An llm-based architecture for adaptive tutoring and social reasoning},
  author={Garello, Luca and Belgiovine, Giulia and Russo, Gabriele and Rea, Francesco and Sciutti, Alessandra},
  booktitle={2025 IEEE/RSJ International Conference on Intelligent Robots and Systems (IROS)},
  pages={9239--9246},
  year={2025},
  organization={IEEE}
}

@inproceedings{aly2015multimodal,
  title={Multimodal adapted robot behavior synthesis within a narrative human-robot interaction},
  author={Aly, Amir and Tapus, Adriana},
  booktitle={2015 IEEE/RSJ International Conference on Intelligent Robots and Systems (IROS)},
  pages={2986--2993},
  year={2015},
  organization={IEEE}
}

@article{wilcoxon1945individual,
  title={Individual comparisons by ranking methods},
  author={Wilcoxon, Frank},
  journal={Biometrics bulletin},
  volume={1},
  number={6},
  pages={80--83},
  year={1945},
  publisher={JSTOR}
}

@article{holm1979simple,
  title={A simple sequentially rejective multiple test procedure},
  author={Holm, Sture},
  journal={Scandinavian journal of statistics},
  pages={65--70},
  year={1979},
  publisher={JSTOR}
}

@article{ahn2022can,
  title={Do as i can, not as i say: Grounding language in robotic affordances},
  author={Ahn, Michael and Brohan, Anthony and Brown, Noah and Chebotar, Yevgen and Cortes, Omar and David, Byron and Finn, Chelsea and Fu, Chuyuan and Gopalakrishnan, Keerthana and Hausman, Karol and others},
  journal={arXiv preprint arXiv:2204.01691},
  year={2022}
}

@article{zhang2023large,
  title={Large language models for human--robot interaction: A review},
  author={Zhang, Ceng and Chen, Junxin and Li, Jiatong and Peng, Yanhong and Mao, Zebing},
  journal={Biomimetic Intelligence and Robotics},
  volume={3},
  number={4},
  pages={100131},
  year={2023},
  publisher={Elsevier}
}

@article{driess2023palm,
  title={Palm-e: An embodied multimodal language model},
  author={Driess, Danny and Xia, Fei and Sajjadi, Mehdi SM and Lynch, Corey and Chowdhery, Aakanksha and Ichter, Brian and Wahid, Ayzaan and Tompson, Jonathan and Vuong, Quan and Yu, Tianhe and others},
  journal={arXiv preprint arXiv:2303.03378},
  year={2023}
}

\end{document}